\documentclass[lettersize,journal]{IEEEtran}
\usepackage{amsmath,amsfonts}
\usepackage{algorithmic}
\usepackage{algorithm}
\usepackage{array}
\usepackage[caption=false,font=normalsize,labelfont=sf,textfont=sf]{subfig}
\usepackage{textcomp}
\usepackage{stfloats}
\usepackage{url}
\usepackage{verbatim}
\usepackage{graphicx}
\usepackage{cite}
\usepackage{amssymb}
\usepackage{multicol}
\usepackage{multirow}
\usepackage{bbm, dsfont}
\usepackage{circledtext}
\usepackage{utfsym}
\usepackage{bbding}
\usepackage{amssymb}
\usepackage[pagebackref,breaklinks,colorlinks]{hyperref}
\begin{document}

\title{EipFormer: Emphasizing Instance Positions in 3D Instance Segmentation}
\author{Mengnan Zhao, Lihe Zhang$^\dagger$, Yuqiu Kong and Baocai Yin}

\markboth{Journal of \LaTeX\ Class Files,~Vol.~14, No.~8, August~2021}%
{Shell \MakeLowercase{\textit{et al.}}: A Sample Article Using IEEEtran.cls for IEEE Journals}


\maketitle

\begin{abstract}
3D instance segmentation plays a crucial role in comprehending 3D scenes. 
Despite recent advancements in this field, existing approaches exhibit certain limitations. 
These methods often rely on fixed instance positions obtained from sampled representative points in vast 3D point clouds, using center prediction or farthest point sampling.
However, these selected positions may deviate from actual instance centers, posing challenges in precisely grouping instances.
Moreover, the common practice of grouping candidate instances from a single type of coordinates introduces difficulties in identifying neighboring instances or incorporating edge points.
To tackle these issues, we present a novel Transformer-based architecture, EipFormer, which comprises progressive aggregation and dual position embedding. 
The progressive aggregation mechanism leverages instance positions to refine instance proposals.
It enhances the initial instance positions through weighted farthest point sampling and further refines the instance positions and proposals using aggregation averaging and center matching.
Additionally, dual position embedding superposes the original and centralized position embeddings, thereby enhancing the model performance in distinguishing adjacent instances. 
Extensive experiments on popular datasets demonstrate that EipFormer achieves superior or comparable performance compared to state-of-the-art approaches.

\end{abstract}


\section{Introduction}
\IEEEPARstart{T}{he} establishment of large-scale 3D datasets \cite{wei2020mitoem,chen2022stpls3d} has significantly advanced the field of 3D scene understanding \cite{hou2021exploring,chen20224dcontrast,jaritz2019multi,zhang2021holistic}. Within this realm, the 3D instance segmentation task \cite{jiang2020end,wen2020cf} occupies a crucial position and serves diverse practical applications such as autonomous driving \cite{zhou2020joint,wang2021solo}, robot navigation \cite{xie2021unseen}, and medical imaging \cite{zanjani2019mask}. The core objective of this task lies in accurately labeling and segmenting individual instances within 3D point clouds \cite{hu2018semantic,li2022joint}.

\begin{figure}[htpb]
    \begin{center}
        
        \includegraphics[width=1\linewidth]{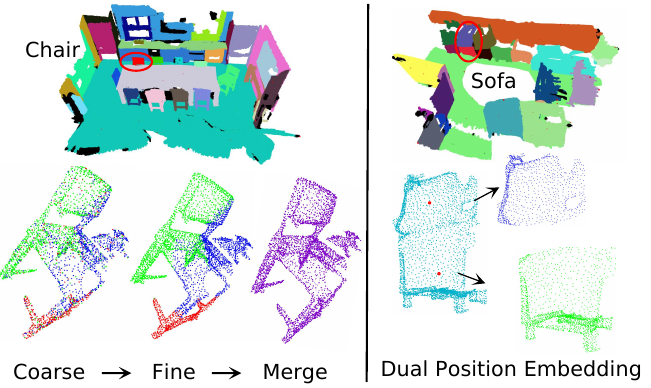}
    \end{center}
    \vspace{-4mm}
    \caption{Intuitions behind the progressive aggregation and dual position embedding. Left: The progressive aggregation is composed of three stages: coarse, fine, and merge.
    Right: The dual position embedding superposes the original and centralized position embeddings, effectively separating neighboring instances while preserving the shape of each individual instance. The red dot represents the ground truth instance center. }
    \label{fig1}
\end{figure}

Existing 3D instance segmentation methods can be categorized into distinct pipelines \cite{Schult23ICRA,chen20223}. Top-down approaches involve generating rough proposals such as bounding boxes or semantic segmentation results, followed by the refinement of these proposals \cite{yang2019learning}. However, top-down methods \cite{liu2020learning} suffer from an issue that errors stemming from earlier proposals propagate into instance grouping. For instance, semantic segmentation misclassifications can induce falsified instance segmentation results \cite{wang2018sgpn,hou20193d}. Consequently, researchers have shown increasing interest in bottom-up techniques \cite{engelmann20203d,chen2021hierarchical}. This pipeline first acquires point-wise representations, such as coordinates or learnable vectors, and then employs a grouping process that clusters points into instances based on these representations \cite{wang2019associatively,jiang2020pointgroup}. In this work, we focus on improving the 3D instance segmentation task in the bottom-up manner.

As previously mentioned, the bottom-up approach can be divided into point-wise representation learning and instance grouping procedures. The former involves two components: appearance representations learned from point colors by a 3D backbone \cite{cciccek20163d, graham20183d}, and position representations derived from either the original point coordinates $\textbf{C}$ \cite{liao2021point} or the centrally shifted point coordinates $\textbf{C+O}$ \cite{engelmann20203d}. Notably, the model based on $\textbf{C}$ faces challenges in distinguishing adjacent instances, while the model with $\textbf{C+O}$ tends to neglect edge points of instances \cite{jiang2020pointgroup}. Hence, PointGroup \cite{jiang2020pointgroup} clusters points within $\textbf{C}$ and $\textbf{C+O}$ to produce candidate instance sets $\textbf{Q}_c$ and $\textbf{Q}_{c+o}$, respectively. which is realized based on a pre-defined radius and the distance between points. In terms of the latter step, researchers typically generate representative points as fixed instance positions and then group points based on these instance positions. Representative points are commonly obtained through center voting \cite{qi2019deep, engelmann20203d} or farthest point sampling \cite{SPFormer}. However, due to numerous points within 3D point clouds, these representative points may deviate seriously from the real instance centers. Consequently, the grouping procedure with imprecise instance positions struggles to aggregate precise instances. To address this issue, recent methods \cite{ngo2023isbnet,Schult23ICRA} resort to expensive post-processing techniques like DBScan to further refine instance segmentation results.

In this work, we introduce EipFormer, a novel Transformer network that effectively utilizes various types of coordinates, yielding more precise instances without relying on post-processing techniques. EipFormer focuses on generating and leveraging instance centers, primarily through progressive aggregation and dual position embedding. The intuitions behind these components are illustrated in Figure \ref{fig1}.

The progressive aggregation aims to aggregate precise instance proposal features from point features across three stages: coarse, fine, and merge.
In the coarse stage, we focus on learning coarse instance proposal features with fixed instance proposal positions. 
To enhance the initial positions, we design a weighted farthest point sampling. Subsequently, the existing proposal matching assigns targets for each instance proposal. Instance proposal features are aggregated based on these positions and targets using self-attention and cross-attention mechanisms of the Transformer.
Moving to the fine stage, we refine both the positions and features of instance proposals. 
The position of each instance proposal is updated with the average coordinate of its corresponding matched points. 
We then introduce center matching as the assignment algorithm that establishes a one-to-one correspondence between instance proposal positions and targets. 
This stage effectively incorporates local contexts into instance proposal features while it may omit several instances.
To address this, the merge stage inherits the refined instance proposals and supplements the omitted instances from the fine stage. This is realized based on the refined instance proposal positions and the proposal matching.

In contrast to PointGroup, EipFormer integrates dual position embedding directly, $i.e.$, combining the original and centralized position embeddings into a unified representation. This choice is made because EipFormer employs a Transformer network, and utilizing attention techniques to cluster instances within sets $\textbf{C}$ and $\textbf{C+O}$ separately would double memory consumption.
The dual position embedding serves to separate adjacent instances while preserving the original shape of individual instances. For instance, consider instances $\textbf{Q}_1$ ($\textbf{C}_1$ $\in$ $[-a, a]$) and $\textbf{Q}_2$ ($\textbf{C}_2$ $\in$ $[a, 2a]$); their new instance scopes become $\textbf{C}_1$ $\in$ $[-a, a]$ and $\textbf{C}_2$ $\in$ $[2.5a, 3.5a]$.

Additionally, given the abundance of points in 3D point clouds, aggregating instance features from all points results in significant memory consumption. To address this, we introduce a simple yet effective class-aware point sampling method, filtering invalid points for each instance.

The contributions are summarized in four aspects:
\begin{itemize}
\setlength{\itemsep}{2pt}
\setlength{\parsep}{2pt}
\setlength{\parskip}{2pt}
\item We propose a progressive aggregation strategy, utilizing weighted farthest point sampling, aggregation averaging, and center matching operations to achieve coarse-to-fine instance feature generation; 
\item To comprehensively distinguish neighboring instances, we present the dual position embedding based on the center offset prediction branch;
\item We devise a class-aware point sampling to reduce memory consumption without greatly sacrificing performance;
\item Extensive experiments demonstrate that our proposed EipFormer achieves favorable performance against the state-of-the-art methods on various instance segmentation datasets. 
Notably, our EipFormer abstains from employing any post-processing techniques.
\end{itemize}

\section{Related Work}\label{sec: 2}
In this section, we begin by providing a comprehensive description of 3D backbone architectures for learning point-wise representations. Subsequently, we introduce various 3D instance segmentation methods.
\subsection{3D Backbone}
Representation learning is extensively studied in 3D scene understanding due to its efficacy in capturing geometric and color features \cite{yin2021bridging,he2023prototype}. There are two common paradigms for 3D representation learning: pixel-based \cite{cheng2021net} and voxel-based \cite{gu20193d} approaches.

For the pixel-based paradigm, PointNet \cite{qi2017pointnet} directly consumed unordered point clouds. 
To capture local structures within point clouds, PointNet++ \cite{qi2017pointnet++} designed a hierarchical neural network. It used farthest-point sampling to select centroids and recursively applied PointNet to extract features from groups centered around these centroids.
PointNet and PointNet++ leveraged the permutation invariant property of 3D point clouds by incorporating a consistent transformation. 
Instead, PointCNN \cite{li2018pointcnn} hierarchically aggregated features of the representative points within their $k$ neighbors using the X-conv layer and learned the transformation function applied to these points.

The voxel-based architectures include extensions of well-established 2D designs into the realm of 3D representations.
For instance, 3D U-Net \cite{cciccek20163d} extended the previous 2D U-Net \cite{ronneberger2015u} architecture by replacing 2D operations with their 3D counterparts. 
Similarly, I3D \cite{carreira2017quo} proposed a 3D ConvNet by inflating a 2D ConvNet, an approach applicable to networks like ResNet \cite{he2016deep} and VGG \cite{simonyan2014very}.
Notably, 3D point clouds are inherently sparse, which challenges the direct application of dense convolutional networks to such sparse data. 
Therefore, in 3D vision tasks such as 3D instance segmentation and 3D object detection \cite{yan2018second},  sparse and submanifold sparse convolutional networks have emerged \cite{graham20183d}.

\begin{figure*}[t]
    \begin{center}
        
        \includegraphics[width=1\linewidth]{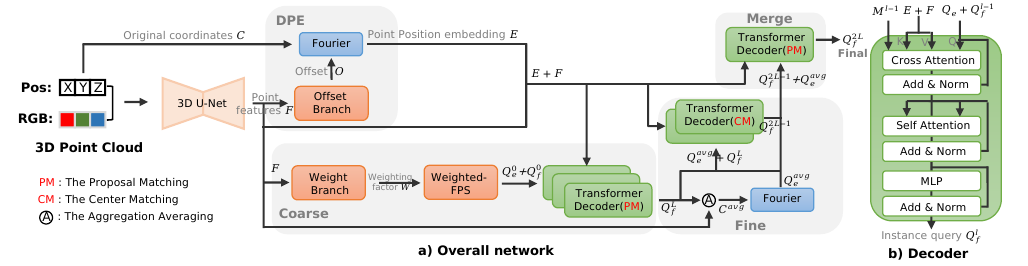}
    \end{center}
    \vspace{-4mm}
    \caption{Illustration of the network architecture. a) The proposed method consists of dual position embedding (DPE) and progressive aggregation.
    The latter consists of the Coarse, Fine and Merge stages.
    Given a 3D point cloud, the 3D U-net extract voxel-wise features. The dual position embedding leverages the original coordinates and the centrally shifted coordinates generated by offset branch. The progressive aggregation generates the initial instance positions using the weighted farthest point sampling and the refined instance positions $\mathbf{C}^{avg}$ by averaging the matched voxels for each aggregated instance. To learn precise instance query features, we utilize the proposal matching in the coarse and merge stages, and the center matching in the fine stage. b) The details of the Transformer decoder.
    $\mathbf{M}^{l-1}$ denotes the instance mask, $\mathbf{M}^{l-1}\in\mathbb{R}^{k\times M}$. $M$ and $k$ are the number of voxels and instance queries, respectively. $\mathbf{M}^{l-1}_{i,j}$ is set to 0 for the positive instance segmentation result and to $-\infty$ for the negative instance segmentation result.
    The instance segmentation result is calculated by the dot product between $\mathbf{Q}_{i,f}^{l-1}$ and $\mathbf{F}_j$.
    Here, $\mathbf{Q}_{i,f}^{l-1}$ is the $i$-th instance query and $\mathbf{F}_j$ is the $j$-th voxel features.
    }
    \label{fig2}
\end{figure*}
\subsection{3D Instance Segmentation}
By utilizing the above mentioned networks \cite{carreira2017quo, qi2017pointnet++} as feature backbones, two popular strategies, top-down and bottom-up, are frequently employed to yield 3D instance segmentation results.
The top-down method generates proposals such as the bounding box or semantic region, and then refines each proposal to yield the instance segmentation results.
For instance, 
3D-SIS \cite{hou20193d} combined the 3D geometry features with 2D RGB features to detect 3D region proposals using RoIPooling.
Contrasting approaches emerge in the establishment of associations between ground truth and predicted instances. GSPN \cite{yi2019gspn} drew inspiration from geometric comprehension and 3D-SIS took cues from 3D anchors like Mask R-CNN. In contrast, 3D-BoNet \cite{yang2019learning} harnessed the existing Hungarian algorithm with the designed bounding box association cost. 
Moreover, GICN \cite{liu2020learning} approximated instance center distributions using Gaussian center heatmaps and utilized filtered center candidates for predicting instance masks.
However, the rough proposals generated by these methods, being potentially imprecise, may hinder the improvement of instance segmentation quality. 

In contrast, bottom-up methods first compute the pointwise embedding and then group points into instances.
Regarding the grouping process, SGPN \cite{wang2018sgpn} predicted pointwise grouping proposals utilizing features derived from PointNet \cite{qi2017pointnet}.
To reduce potential group proposals, 3D-MPA \cite{engelmann20203d} adopted an object-centric strategy to acquire representative points.
As the clustering step in 3D-MPA consumes positive proposals but neglects wrongly classified proposal fragments, HAIS \cite{chen2021hierarchical}
performed set grouping after point grouping using predefined clustering bandwidths.
However, these bandwidths require statistical information about instance sizes and vary across various datasets.
Moreover, Mask3D \cite{Schult23ICRA} and SPFormer \cite{SPFormer} employed the farthest point sampling to produce fixed instance positions that may deviate from the precise instance centers.
In comparison, our progressive aggregation not only enhances the quality of instance positions through weighted farthest point sampling and aggregation averaging but also effectively learns instance appearance features with center matching.

As for the point representation learning process, ASIS \cite{wang2019associatively} learned the semantic-aware instance embeddings by attracting point representations of the same instance towards their mean value and encouraging those of different instances to repel each other.
JSNet \cite{zhao2020jsnet} similarly integrated features from different backbone layers and also aggregated semantic features to facilitate instance segmentation.
However, our experiments demonstrate that the semantic prediction actually degrades the instance segmentation performance.
Moreover, Mask3D and SSTNet \cite{liang2021instance} utilized the original and centrally shifted coordinates to describe instance positions, respectively.
Furthermore, while PointGroup \cite{jiang2020pointgroup} leveraged the strengths of the original and centralized geometric information, it separately grouped instances from two distinct proposal sets.
In contrast, our approach directly superposes the original and centralized position embeddings.

There are also approaches incorporating bottom-up and up-down techniques.
ISBNet \cite{ngo2023isbnet} combined 3D bounding boxes with dynamic convolutions.
Unlike hard grouping in 3D-MPA and HAIS, SoftGroup \cite{vu2022softgroup} and SoftGroup++ \cite{vu2022softgroup++} employed soft grouping to mitigate the problems caused by semantic prediction errors.
Compared with these state-of-the-art methods, our work realizes superior or comparable performance. 

\section{Methods}\label{section3}
The overall architecture of our EipFormer is visualized in Figure \ref{fig2}. It comprises a sparse convolutional U-net \cite{choy20194d} as the feature backbone, a dual position embedding module for constructing voxel-wise features, and a progressive aggregation process for gathering instance query features from voxel features. The instance segmentation results are determined based on the dot product between the instance query features and the voxel features.

\subsection{Dual Position Embedding}
Given a 3D point cloud $\mathbf{P}\in \mathbb{R}^{N\times 6}$ with $N$ points as input, in which each point is characterized by its 3D coordinate and RGB color, we voxelize $\mathbf{P}$ and utilize a 3D sparse U-net backbone to learn voxel-wise features $\mathbf{F}\in \mathbb{R}^{{M}\times {D}}$ from the terminal layer. 
The coordinates of voxels are denoted as $\mathbf{C}\in \mathbb{R}^{{M}\times {3}}$.
Here, $M$ signifies the number of voxels and $D$ denotes the dimension of extracted features.

Considering the difficulty of distinguishing adjacent instances and integrating edge points of instances, we follow \cite{jiang2020pointgroup} to utilize both the original and centrally shifted voxel coordinates.
Instead of separately extracting candidate instances from $\mathbf{C}$ and $\mathbf{C+O}$, we superpose representations of them,
\begin{equation}\label{eq3} 
\mathbf{E} = \textup{Fourier}(\mathbf{C}) + \textup{Fourier}(\mathbf{C} + \mathbf{O}),
\end{equation}
where $\textup{Fourier}(\cdot)$ indicates the Fourier position encoding \cite{tancik2020fourier}.
$\mathbf{O}$ represents the predicted voxel-wise offsets, where $\mathbf{O}\in\mathbb{R}^{M\times 3}$.
We employ a center offset prediction branch to predict $\mathbf{O}$ based on the voxel features $\mathbf{F}$ and coordinates $\mathbf{C}$.
This branch consists of a two-layer MLP.
$\mathbf{O}$ is supervised by the center regression loss $\mathcal{L}_{reg}$ and the direction loss $\mathcal{L}_{dir}$ \cite{jiang2020pointgroup},
\begin{equation}\label{eq1}
    \mathcal{L}_{reg} = \frac{1}{\sum_{i=1}^{M}\mathds{1}_{(i)}}\cdot\sum_{i=1}^{{M}}\mathds{1}_{(i)}\|o_i - o_i^*\|_2,
\end{equation}
\begin{equation}\label{eq2} 
    \mathcal{L}_{dir} = - \frac{1}{\sum_{i=1}^{M}\mathds{1}_{(i)}} \cdot\sum_{i=1}^{{M}} \mathds{1}_{(i)}\frac{o_i}{\|o_i\|_2}\cdot\frac{o_i^*}{\|o_i^*\|_2},
\end{equation}
where $\mathds{1}_{(i)}$ signifies whether the $i$-th voxel belongs to any instance. 
$o_i^*$ represents the ground truth center offset of $o_i$, $o_i\in\mathbf{O}$. 
$\|\cdot\|_2$ means the $2$-norm function. 
As depicted in Figure \ref{fig1}, the dual position embedding aims to separate adjacent instances to distinct positions.

\subsection{Progressive Aggregation}\label{3.2}
Upon obtaining the voxel-wise features $\mathbf{F}$ and position embeddings $\mathbf{E}$, researchers typically apply farthest point sampling to select $k$ representative points as initial instance queries $\mathbf{Q}^0$ = $\{\mathbf{Q}_f^0, \mathbf{Q}_e^0\}$, where $\mathbf{Q}_f^0\in\mathbf{F}$ and $\mathbf{Q}_e^0\in\mathbf{E}$.
Subsequently, the point features are aggregated into instance query features $\mathbf{Q}_f$ based on the fixed position embedding $\mathbf{Q}_e$, where $\mathbf{Q}_e$ = $\mathbf{Q}_e^0$.
Notably, the 3D point clouds typically contain numerous points. 
Hence, the sampled point positions may seriously deviate from the real instance centers, which poses a challenge in aggregating precise $\mathbf{Q}_f$.
In the following, we elaborate the proposed progressive aggregation, which aims to mitigate this issue through three stages: coarse, fine, and merge.

\subsubsection{Coarse}\label{section3.2.1}
This stage will produce coarse instance query features.
To generate initial instance positions, we present a weighted farthest point sampling strategy.
The learnable weight $\mathbf{W}\in\mathbb{R}^{M}$ acts as a distance weighting factor during sampling and is supervised by ground truth center offsets,
\begin{equation}\label{eq4} 
\mathcal{L}_{fore} = \frac{1}{\sum_{i=1}^{M}\mathds{1}_{(i)}}\cdot\sum_{i=1}^{{M}}\mathds{1}_{(i)}|w_i - e^{-\alpha\cdot\|o_i^*\|_2}|,
\end{equation}
\begin{equation}\label{eq5} 
\mathcal{L}_{back} = \frac{1}{\sum_{i=1}^{M}\hat{\mathds{1}_{(i)}}}\cdot\sum_{i=1}^{{M}}\hat{\mathds{1}_{(i)}}|w_i|.
\end{equation}
The foreground (indicated by $\mathds{1}_{(i)}$) and background (indicated by $\hat{\mathds{1}_{(i)}}$) voxels are constrained by losses $\mathcal{L}_{fore}$ and $\mathcal{L}_{back}$, respectively. 
Here, $w_i$ represents the weight for the $i$-th voxel and $w_i\in \mathbf{W}$.
The hyperparameter $\alpha$ controls the average weight to be around 0.2. 
$e$ is the Euler number and $|\cdot|$ denotes the absolute operation.
In this way, the initially sampled points closely approximate instance centers.

After obtaining initial instance queries $\mathbf{Q}^0$, we progressively update instance query features $\mathbf{Q}_f$ with the existing Transformer network and the fixed instance query positions $\mathbf{Q}_e^0$.
The structure of the Transformer decoder is displayed in Figure \ref{fig2}.
Specifically, the query features undergo refinement through the cross-attention that involves attending to positive voxel-wise features, and the self-attention that enables interactions among different instance queries,
\begin{equation}\label{eq6}
\mathbf{Q}_f^{l^\prime} = \textup{softmax}(\frac{(\mathbf{Q}_f^{l-1} + \mathbf{Q}_e^0)(\mathbf{F} + \mathbf{E}) + \mathbf{M}^{l-1}}{\sqrt{{D}}})\mathbf{F},
\end{equation}
\begin{equation}\label{eq7} 
\mathbf{Q}_f^{l} = \textup{softmax}(\frac{(\mathbf{Q}_f^{l^\prime} + \mathbf{Q}_e^0)(\mathbf{Q}_f^{l^\prime} + \mathbf{Q}_e^0)}{\sqrt{{D}}})\mathbf{Q}_f^{l^\prime},
\end{equation}
where $l$ represents the $l$-th Transformer decoder layer, $l\in[1, L]$.
$\mathbf{M}^{l-1}$ denotes the instance mask, $\mathbf{M}^{l-1}\in\mathbb{R}^{k\times M}$. $M$ and $k$ are the number of voxels and instance queries, respectively.  $\mathbf{M}^{l-1}_{i,j}$ is set to 0 for the positive instance segmentation result and to $-\infty$ for the negative instance segmentation result.
The instance segmentation result is calculated by the dot product between $\mathbf{Q}_{i,f}^{l-1}$ and $\mathbf{F}_j$.
Here, $\mathbf{Q}_{i,f}^{l-1}$ is the $i$-th instance query and $\mathbf{F}_j$ is the $j$-th voxel features.
$D$ denotes the dimension of extracted features.

In each Transformer decoder layer, we follow \cite{Schult23ICRA} to reassign targets (or ground-truth instances $\mathbf{I}^*$) for the predicted instance segmentation results $\mathbf{I}\in\mathbb{R}^{k\times(M+S)}$, denoted as proposal matching.
Here, $S$ is the number of semantic categories, and $\mathbb{R}^{k\times M}$ represents the shape of binary instance masks.
The proposal matching is accomplished using the Hungarian algorithm along with the following association cost,
\begin{equation}\label{eq8} 
\mathcal{L}_{ac} = \lambda_1\mathcal{L}_{dice}(\mathbf{I}, \mathbf{I}^*) + \lambda_2\mathcal{L}_{mask}(\mathbf{I}, \mathbf{I}^*) + \lambda_3\mathcal{L}_{seg}(\mathbf{I}, \mathbf{I}^*).
\end{equation}
Eq. (\ref{eq8}) consists of several components, each with its respective weight $\lambda_{1\sim 3}$. The dice loss $\mathcal{L}_{dice}$ \cite{deng2018learning} quantifies the similarity between instance masks, addressing the issue of imbalance between positive and negative samples. Additionally, $\mathcal{L}_{mask}$ represents a binary cross-entropy loss that distinguishes between foreground and background voxels, while $\mathcal{L}_{seg}$ is a standard cross-entropy loss used to generate semantic labels for the predicted instances.
Importantly, instead of optimizing network parameters, the association cost in the proposal matching process is used for for assigning regression targets.

\subsubsection{Fine}\label{section3.2.2}
In the above stage, we obtain coarse instance query features $\mathbf{Q}_f^{L}$ based on the fixed instance query positions $\mathbf{Q}_e^0$.
Combined with the experimental phenomenon that the targets assigned by the proposal matching for $\textbf{I}$ vary across decoder layers, it becomes evident that each instance query position does not uniquely describe a specific target.
Next, we further refine both instance query features and positions, aiming to establish a one-to-one correspondence between instance positions and instance targets.

To update the position embeddings of instance queries, we introduce the aggregation averaging. 
This is achieved by averaging the voxel coordinates within each predicted instance.
\begin{equation}\label{eq9} 
\mathbf{C}^{avg}_i = \frac{\sum \mathbf{C}[\mathbf{Q}_{i,f}^{L}\mathbf{F}>0]}{\sum [\mathbf{Q}_{i,f}^{L}\mathbf{F}>0]},
\end{equation}
\begin{equation}\label{eq92} 
\mathbf{Q}_e^{avg} = \textup{Fourier}(\mathbf{C}^{avg}),
\end{equation}
$\mathbf{C}[\mathbf{Q}_{i,f}^{L}\mathbf{F}>0]$ locates matched voxel coordinates for the $i$-th instance query.
In terms of refining instance query features, the center matching is designed to assign the targets for the predicted instances with ground-truth instances.
The association cost $\mathcal{L}_{cm}$ takes the refined instance positions $\mathbf{C}^{avg}$ and groung-truth instance centers $\mathbf{C}+\mathbf{O}^*$ as inputs, 
\begin{equation}\label{eq10} 
\mathcal{L}_{cm}(\mathbf{C}^{avg}, \mathbf{C}+\mathbf{O}^*),
\end{equation}
where $\mathcal{L}_{cm}$ is a 2-norm function and calculates the distance between $\mathbf{C}^{avg}$ and $\mathbf{C}+\mathbf{O}^*$. $\mathbf{O}^*$ denotes the ground truth offsets of voxels to their respective centers. 
With the refined instance positions and assigned targets, we apply additional $L-1$ Transformer decoder layers to further refine the features of instance queries.
Notably, for all decoder layers in this stage, we only calculate the instance positions in Eq. (\ref{eq92}) and targets in Eq. (\ref{eq10}) once and fix them.
This ensures that each instance position uniquely describe a specific target across various Transformer decoder layers.
Similarly, the association cost in center matching is used for assigning targets rather than optimizing model weights.

\subsubsection{Merge}\label{section3.2.3}
The Fine stage incorporates local context into instance query features, specifically by focusing on points close to instance centers. However, this approach potentially omits several instances or produces fragmented instances. To address this, we introduce a Merge stage. The term $\mathbf{Q}_e^{avg}$ in Eq. (\ref{eq92}) is also employed as the instance position embedding to maintain consistency with the Fine stage, allowing the Merge stage to inherit the refined instance proposals. For the assignment of targets to predicted instances, we opt for proposal matching. Through experiments, we observe that this matching algorithm tends to adjust the targets for omitted or fragmented instances instead of modifying the matched targets for the top refined instance proposals.
Based on $\mathbf{Q}_e^{avg}$ and the proposal matching, the Merge stage is implemented by a single Transformer decoder layer.

\subsection{Class-Aware Point Sampling}\label{section3.3}
The preceding section explains how to aggregate point features to update instance query features. 
However, owing to irregular distribution and numerous points within 3D point clouds, this aggregation process significantly consumes memory. 
Meanwhile, during the cross-attending process, ensuring adequate information for updating instance query features with randomly sampled voxels presents a challenge.
This is mainly because the majority of sampled points originate from backgrounds or belong to large instances.
To address this issue, we introduce a class-aware point sampling strategy to carefully select voxels for the training process.
Specifically, we first choose $n$ voxels for each instance and then randomly select additional voxels from the remaining point cloud. Furthermore, to prevent overfitting to tiny instances, we impose a restriction on the number of selected voxels per instance, ensuring it remains below half of the total points in the instance.

\subsection{Loss and Confidence Calculation}
In Sec. $\ref{3.2}$, we build the correspondence between instances predicted by each Transformer decoder layer and ground truth instances. 
We then employ the following loss function to optimize each layer,
\begin{equation}\label{eq11} 
\mathcal{L}_{all} = \mathcal{L}_{reg} + \mathcal{L}_{dir}  + \mathcal{L}_{fore} + \mathcal{L}_{back} + \mathcal{L}_{ac},
\end{equation}
where $\mathcal{L}_{reg}$ and $\mathcal{L}_{dir}$ constrain point-wise offsets, 
$\mathcal{L}_{fore}$ and $\mathcal{L}_{back}$ learn the weighting factors of weighted farthest point sampling.
The default balance weights for these losses are all set to 1, without manual fine-tuning.
Additionally, we borrow $\mathcal{L}_{ac}$ from \cite{Schult23ICRA} to update instance query features.
Overall, these losses can be balanced easily.

Following \cite{cheng2022masked}, the prediction confidence of each instance is quantified by multiplying the dominant semantic confidence with the mean mask confidence. The mean mask confidence for the $i$-th instance is expressed as
\begin{equation}\label{eq12} 
\frac{\sum\sigma(\mathbf{Q}_{i,f}^{l}\mathbf{F})\cdot[\mathbf{Q}_{i,f}^{l}\mathbf{F}>0]}{\sum [\mathbf{Q}_{i,f}^{l}\mathbf{F}>0]}.
\end{equation}

\section{Experiments}
Our experiments are divided into four sections. In Section \ref{section4.1}, we outline the experimental settings, providing details on dataset specifications, evaluation metrics, and comparison methods. Section \ref{section4.2} presents quantitative results on 3D instance segmentation methods across various datasets. Additionally, visual examples and failure cases are showcased in Section \ref{section4.3}. In Section \ref{section4.4}, we present the results of ablation studies.

\subsection{Experimental Settings} \label{section4.1}
\subsubsection{Datasets} We evaluate EipFormer on several popular instance segmentation datasets: STPLS3D \cite{chen2022stpls3d}, S3DIS \cite{armeni20163d}, and ScanNetV2 \cite{dai2017scannet}.
\begin{itemize}
\setlength{\itemsep}{2pt}
\setlength{\parsep}{2pt}
\setlength{\parskip}{2pt}
\item[-] STPLS3D \cite{chen2022stpls3d}: A mix of real-world and synthetic environments with 25 urban scenes densely annotated with 14 instance classes. Following \cite{vu2022softgroup,Schult23ICRA}, we split this dataset into training and validation sets. Scenes numbered 5, 10, 15, 20, and 25 are allocated as the validation set, while the remaining scenes are designated for the training set.
\item[-] S3DIS \cite{armeni20163d}: Covers 271 indoor scenes from 6 distinct areas, annotated with instance masks among 13 categories. To comprehensively assess model performance for each area, our approach involves training models using five of these areas and subsequently evaluating them on the remaining unexplored area. In the context of a 6-fold cross-evaluation setting, we compute the average performance across all six areas.
\item[-] ScanNetV2 \cite{dai2017scannet}: Consists of 1201, 312, and 100 scans for training, validation, and testing, respectively, with a total of 18 object categories. Both the experimental results on the validation and hidden test sets are reported. For this dataset, we adopt a memory-efficient strategy to generate instance segmentation results, involving the computation of dot products between instance queries and aggregated point features within segments \cite{zhao2022divide,nekrasov2021mix3d}.
\end{itemize}

\subsubsection{Evaluation metrics}
For all datasets, we employ mean average precision (AP), AP$_{50}$, and AP$_{25}$ as evaluation metrics \cite{han2020occuseg,lahoud20193d,chu2022twist}. The AP metric averages scores over a range of Intersection over Union (IoU) thresholds, ranging from 50\% to 95\%, with a stride of 5\%. Additionally, AP$_{50}$ and AP$_{25}$ represent scores with IoU thresholds of 50\% and 25\%, respectively.
Furthermore, on the S3DIS dataset, we report mean precision (mPrec$_{50}$) and mean recall (mRec$_{50}$) with an IoU threshold of 50\%, as defined in \cite{hui2022learning,dong2022learning}.

\subsubsection{Baselines}
We include advanced methods as baselines, $i.e.$, GSPN \cite{yi2019gspn}, PointInst3D \cite{he2022pointinst3d}, 3D-MPA \cite{engelmann20203d}, PointGroup \cite{jiang2020pointgroup}, SSTNet \cite{liang2021instance}, HAIS \cite{chen2021hierarchical}, SoftGroup \cite{vu2022softgroup}, Di\&Co3D \cite{zhao2022divide}, DKNet \cite{wu20223d}, ISBNet \cite{ngo2023isbnet}, and Mask3D \cite{Schult23ICRA}.

\subsubsection{Implementation details}
Without special statement, we follow \cite{Schult23ICRA} to employ the Minkowski Res16UNet34C \cite{choy20194d} as the feature backbone.
The models are trained for 600 epochs on ScanNetV2 and STPLS3D, and 1000 epochs on S3DIS. 
Our optimization process employs the AdamW optimizer and the one-cycle scheduler with a maximal learning rate of 2$\times$10e-4.
Voxel sizes are set to 0.333m, 0.04m, and 0.02m for STPLS3D, S3DIS, and ScanNetV2 respectively.
Data augmentation incorporates the procedures outlined in Mask3D.
For a balance between performance and memory consumption, we set the number of queries ($k$) and sampled points to 100 and 12800, respectively. 
$n$ in the class-aware point sampling is set to 128, which is equal to the ratio between the sampling points and the number of instance queries.
With a batch size of 5, the average running time for ScanNetV2 is approximately 80 hours when executed on a single GeForce RTX 3090 GPU.
Throughout the training process, model performance is assessed on the validation set at regular intervals of 10 epochs.

For other hyper-parameters, we set $\lambda_{1}$, $\lambda_{2}$, and $\lambda_{3}$ in Eq. (\ref{eq8}) to 2, 5, and 2, respectively.
Additionally, we empirically fix $L$ in Eq. (\ref{eq9}) at 6 to maintain a comparable model size to prior methods. 
$\alpha$ in Eq. (\ref{eq4}) is set to 25.

\subsubsection{Model sizes}
Compared to the reference method, our network utilizes fewer parameters, as shown in Table \ref{tab2}. Specifically, unlike Mask3D, which employs varying weights for different Transformer decoder layers, our method implements weight sharing across various decoder layers.

\subsubsection{Other details}
While post-processing techniques have shown the potential to enhance instance segmentation performance, they often introduce notable time delays during the evaluation phase. To ensure practicality and efficiency for real-world applications, all our results are reported without using any post-processing techniques. Notably, we do not change the competitors except for Mask3D. For Mask3D, we provide both results with and without using DBScan.

\begin{table}[tpb]
 	\caption{3D instance segmentation results on STPLS3D.
 $^*$ indicates evaluations without using the post-processing.}
 \begin{center}
	\begin{tabular}{llccc } 
		\hline 
        {\bf Methods}&{\bf Venue}&{\bf AP}& {\bf AP}$_{50}$ & {\bf AP}$_{25}$ \\
		\hline
        PointGroup \cite{jiang2020pointgroup}&CVPR 20&23.3&38.5&48.6\\
        HAIS \cite{chen2021hierarchical}&ICCV 21&35.1&46.7&52.8\\
        SoftGroup \cite{vu2022softgroup}&CVPR 22&46.2&61.8&69.4\\
        ISBNet \cite{ngo2023isbnet}&CVPR 23&49.2&64.0&-\\
        Mask3D \cite{Schult23ICRA}&ICRA 23&57.3&74.3&81.6\\
        \hline
        Mask3D$^*$ \cite{Schult23ICRA}&ICRA 23&56.8&73.7&80.8\\
        {\bf Ours$^*$}&-&{\bf 58.9}&{\bf 75.8}&{\bf 83.3}\\
		\hline
	\end{tabular}
 \end{center}
 \label{tab1}
 \vspace{-4mm}
\end{table}

\begin{table}[t]
    \centering
\caption{Model sizes.}
    \begin{tabular}{c|cc}
    \hline 
    {\bf Backbone/Other}& {\bf Mask3D}&{\bf Ours}\\
    \hline
       Params.& 37.856M/1.761M &37.856M/{\bf 1.426M}\\
          \hline
    \end{tabular}
    \label{tab2}
\end{table}

\begin{table}[htpb]
\caption{3D instance segmentation results on S3DIS (Voxel size: 0.04m). Mask3D is the state-of-the-art method on S3DIS.}
 \begin{center}
 	\tabcolsep = 0.14cm
	\begin{tabular}{cl|ccccc } 
		\hline 
        {\bf Area}& {\bf Methods} &{\bf AP} & {\bf AP}$_{50}$&{\bf AP}$_{25}$& {\bf mPrec}$_{50}$&{\bf mRec}$_{50}$\\
		\hline
        \multirow{2}{*}{1}
        & Mask3D \cite{Schult23ICRA}&{\bf 62.92}&{\bf 74.82}&78.77&79.45&{\bf 73.09}\\
        &{\bf Ours}&62.49&73.78&{\bf 80.97}&{\bf 79.46}&71.03\\
        \hline
        \multirow{2}{*}{2}
        & Mask3D \cite{Schult23ICRA}&35.34&45.82&61.81&64.51&{\bf 47.63}\\
        &{\bf Ours}&{\bf 36.78}&{\bf 49.54}&{\bf 63.86}&{\bf 65.66}&46.47\\
        \hline
        \multirow{2}{*}{3}
        & Mask3D \cite{Schult23ICRA}&69.02&81.48&87.16&78.28&74.31\\
        &{\bf Ours}&{\bf 70.48}&{\bf 83.38}&{\bf 87.99}&{\bf 81.05}&{\bf 80.72}\\
        \hline
        \multirow{2}{*}{4}
        & Mask3D \cite{Schult23ICRA}&51.77&65.28&75.44&75.90&61.63\\
        &{\bf Ours}&{\bf 59.60}&{\bf 74.52}&{\bf 79.99}&{\bf 80.49}&{\bf 64.79}\\
        \hline
        \multirow{2}{*}{5}
        & Mask3D \cite{Schult23ICRA}&48.48&61.71&70.07&68.50&58.89\\
        &{\bf Ours}&{\bf 56.60}&{\bf 68.48}&{\bf 73.81}&{\bf 71.96}&{\bf 60.51}\\
        \hline
        \multirow{2}{*}{6}
        & Mask3D \cite{Schult23ICRA}&63.96&73.58&79.82&79.32&71.16\\
        &{\bf Ours}&{\bf 65.03}&{\bf 75.32}&{\bf 83.94}&{\bf 81.24}&{\bf 72.53}\\
        \hline
        \multirow{2}{*}{6-fold}
        &Mask3D \cite{Schult23ICRA}&55.25&67.12&75.51&74.33&64.45\\
        &{\bf Ours}&{\bf 58.50}&{\bf 70.84}&{\bf 78.43}&{\bf 76.64}&{\bf 66.01}\\
		\hline
	\end{tabular}
 \end{center}
	\label{tab3}
\end{table}

\begin{table}[htpb]
\caption{3D instance segmentation results of the 6-
fold cross-evaluation setting on S3DIS.}
 \begin{center}
 	\tabcolsep = 0.13cm
	\begin{tabular}{cl|ccccc } 
		\hline 
        {\bf Voxel size}&{\bf Methods} &{\bf AP} & {\bf AP}$_{50}$&{\bf AP}$_{25}$& {\bf mPrec}$_{50}$&{\bf mRec}$_{50}$\\
        \hline
        \multirow{6}{*}{0.02m}
        &3D-BoNet \cite{yang2019learning} &47.6 &65.6&-&-&-\\
        &3D-MPA \cite{engelmann20203d}&-&-&-&66.7&64.1\\
        &PointGroup \cite{jiang2020pointgroup}&-&64.0&-&69.6&69.2\\
        &DKNet \cite{wu20223d}&-&75.3&71.1&70.3&72.8\\
        &SSTNet \cite{liang2021instance}&54.1&67.8&-&73.5&73.4\\
        &Mask3D \cite{Schult23ICRA}&64.5&75.5&-&72.8&74.5\\
        \hline
        \multirow{2}{*}{0.04m}&Mask3D \cite{Schult23ICRA}&55.3&67.1&75.5&74.3&64.5\\
        &{\bf Ours}&{58.5}&{70.8}&{78.4}&{76.6}&{66.0}\\
		\hline
	\end{tabular}
 \end{center}
	\label{tab4}
\end{table}

\subsection{Benchmark Results}  \label{section4.2}
\subsubsection{STPLS3D}
We first conduct experiments on the recent 3D instance segmentation dataset, STPLS3D, utilizing the Minkowski Res16UNet18B as the feature backbone.
The results are evaluated on the validation set and presented in Table \ref{tab1}.
It can be observed that our method outperforms Mask3D omitting the post-processing technique. Specifically, the proposed EipFormer demonstrates improvements of +2.1/2.1/2.5 in terms of AP/AP$_{50}$/AP$_{25}$ metrics.
Additionally, in comparison to other advanced methods, Table \ref{tab1} indicates the superior performance of our approach.

\subsubsection{S3DIS} 
Since we utilize the same backbone as previous methods to extract voxel-wise features, our method still faces challenges in balancing various point clouds, with the maximum batch size being determined based on the point cloud with the highest number of points. This issue becomes particularly noticeable in datasets with substantial variations in point cloud sizes, as observed in S3DIS.
Therefore, we conduct experiments on the S3DIS dataset with a voxel size of 0.04m due to hardware constraints.

Table \ref{tab3} shows the experimental results across diverse evaluation scenarios of S3DIS.
The proposed EipFormer outperforms the baseline, particularly evident in the AP metrics.
For instance, in Area 5, EipFormer achieves remarkable scores of 56.6/68.5/73.8 in AP/AP$_{50}$/AP$_{25}$, reflecting an enhancement of +8.1/6.8/3.7 compared to the baseline.
Furthermore, our method realizes optimal performance in mPrec$_{50}$/mRec$_{50}$ metrics across the majority of distinct areas. For example, when compared to Mask3D, EipFormer improves performance by 4.59/3.16 in mPrec$_{50}$/mRec$_{50}$ metrics on Area 4.
In addition, we conduct comparative experiments between our method using a voxel size of 0.04m and other state-of-the-art methods employing a voxel size of 0.02m.
The results in Table \ref{tab4} show that the proposed EipFormer still achieves optimal performance on the metric mPrec$_{50}$.

\begin{table*}[htpb]
	\tabcolsep = 0.12cm
	\caption{3D instance segmentation results on the hidden test set of ScanNetV2 in terms of the AP$_{25}$ metric. $^*$ indicates evaluations conducted without utilizing post-processing techniques.}
    \begin{center}
	\begin{tabular}{l|c|ccc ccccc ccccc ccccc} 
		\hline
        &&\multirow{3}{*}{\bf \rotatebox{60}{bathtub}}&\multirow{3}{*}{\bf \rotatebox{60}{bed}}&\multirow{3}{*}{\bf \rotatebox{60}{bkshelf}}&\multirow{3}{*}{\rotatebox{60}{\bf cabinet}}&\multirow{3}{*}{\bf \rotatebox{60}{chair}}&\multirow{3}{*}{\bf \rotatebox{60}{counter}}&\multirow{3}{*}{\bf \rotatebox{60}{curtain}}&\multirow{3}{*}{\bf \rotatebox{60}{desk}}&\multirow{3}{*}{\bf \rotatebox{60}{door}}&\multirow{3}{*}{\bf \rotatebox{60}{other}}&\multirow{3}{*}{\bf \rotatebox{60}{picture}}&\multirow{3}{*}{\bf \rotatebox{60}{fridge}}&\multirow{3}{*}{\bf \rotatebox{60}{s. cur}}&\multirow{3}{*}{\bf \rotatebox{60}{sink}}&\multirow{3}{*}{\bf \rotatebox{60}{sofa}}&\multirow{3}{*}{\bf \rotatebox{60}{table}}&\multirow{3}{*}{\bf \rotatebox{60}{toilet}}&\multirow{3}{*}{\bf \rotatebox{60}{window}}\\
        {\bf Methods}&{\bf AP}$_{25}$\\
        &\\
        \hline

DKNet \cite{wu20223d}&81.5 &100 &93.0 &84.4 &76.5 &91.5 &53.4 &80.5 &80.5 &80.7&65.4 &76.3 &65.0 &100 &79.4 &88.1 &76.6 &100 &75.8\\ 
SoftGroup \cite{vu2022softgroup}&86.5 &100 &96.9&{\bf 86.0} &86.0 &91.3 &55.8 &{\bf 89.9} &91.1 &{\bf 76.0} &82.8 &73.6 &80.2 &98.1 &91.9 &87.5 &87.7&100 &82.0\\
ISBNet \cite{ngo2023isbnet}&84.5&100&97.6 &79.8&79.4&91.6&75.7&66.7&88.2&84.2 &71.5&75.7&83.2&100&90.5&80.3&84.3&100&71.5\\
Mask3D \cite{Schult23ICRA}&{\bf 87.0}&100&98.5&78.2&81.8&93.8&76.0&74.9&92.3&87.7&76.0&78.5 &82.0&100&91.2&86.4&87.8&98.3&82.5\\
\hline

Mask3D$^*$ \cite{Schult23ICRA}&85.2&100&{\bf 98.8}&76.3&81.5	&90.7&66.3&70.9	&91.8&81.1&74.4	&74.7&82.4&100&	91.1&88.0&85.0&100&80.1
\\
{\bf Ours$^*$}&86.5&{\bf 100}&91.2&82.7&{\bf 86.6}&{\bf 94.6}&62.8&82.4&{\bf 96.0}&{\bf 88.3}&73.6&73.1&79.2&{\bf 100}&	{\bf 94.8}&{\bf 91.5}&84.6&94.4&81.0\\
\hline
 \end{tabular}
     \end{center}
	\label{tab6}
 \vspace{-2mm}
\end{table*}

\begin{table}[htpb]
\caption{3D instance segmentation results on ScanNetV2.
 $^*$ indicates evaluations without using the post-processing. 
}
 \begin{center}
	\begin{tabular}{llcccc } 
		\hline 
        \multirow{2}{*}{{\bf Methods}}&\multirow{2}{*}{\bf Venue}& \multicolumn{2}{c}{{\bf ScanNet Val}}& \multicolumn{2}{c}{{\bf ScanNet Test}}\\
        \cline{3-6}
        & &{\bf AP}& {\bf AP}$_{50}$&{\bf AP}& {\bf AP}$_{50}$\\
		\hline
        GSPN \cite{yi2019gspn}&CVPR 19&19.3&37.8&-&30.6\\
        3D-MPA \cite{engelmann20203d}&CVPR 20&35.5&59.1&39.5&-\\
        PointGroup \cite{jiang2020pointgroup}&CVPR 20 &34.8&51.7&40.7&63.6\\
        SSTNet \cite{liang2021instance}&ICCV 21&49.4&64.3&50.6&69.8\\
        HAIS \cite{chen2021hierarchical}&ICCV 21&43.5&64.1&45.7&69.6\\
        DKNet \cite{wu20223d}&ECCV 22&50.8&66.7&53.2&71.8\\
        Di\&Co3D \cite{zhao2022divide}&ECCV 22&47.7&67.2&47.7&70.0\\
        SoftGroup \cite{vu2022softgroup}&CVPR 22&46.0&67.6&50.4&76.1\\
        ISBNet \cite{ngo2023isbnet}&CVPR 23&54.5&73.1&55.9&76.3\\
        Mask3D \cite{Schult23ICRA}&ICRA 23&55.1&73.7&{\bf 56.6}&{\bf 78.0}\\
        \hline
        Mask3D$^*$ \cite{Schult23ICRA}&ICRA 23&53.6&72.5&56.0&76.7\\
        {\bf Ours$^*$}&-&{\bf 56.9}&{\bf 74.6}&56.0&76.1\\
		\hline
	\end{tabular}
 \end{center}
	\label{tab5}
\end{table}

\subsubsection{ScanNetV2}
In this part, we present evaluation results for both the validation and hidden test sets. To assess the model performance on the validation set, we exclusively optimize model parameters using the training data. 
In contrast, for evaluating models on the test set, we combine both the training and validation data during the training phase. 
The instance segmentation results generated on the hidden test set are submitted to the server for evaluation every two weeks.

Table \ref{tab5} presents a comparative analysis on ScanNetV2 between our EipFormer and a range of baselines. 
In comparison with Mask3D, which removes post-processing techniques, our method achieves optimal performance on the validation dataset. 
For instance, the proposed EipFormer demonstrates a performance improvement of +3.3/2.1 on AP/AP$_{50}$ metrics.
Simultaneously, our approach realizes comparable performance to this baseline on the test set. Furthermore, in comparison with other state-of-the-art methods, EipFormer still demonstrates superior performance on the validation dataset.

Table \ref{tab6} presents the class-wise AP$_{25}$ results for 18 classes of ScanNetV2. Notably, our method attains the highest scores in 8 of these classes. Overall, EipFormer shows competitive performance against existing approaches on the test set.

\begin{table}[tpb]
 	\caption{3D object detection results on ScanNetV2.
 $^*$ indicates evaluations without using the post-processing.}
 \begin{center}
	\begin{tabular}{llcc } 
		\hline 
        {\bf Methods}&{\bf Venue}&{\bf AP}& {\bf AP}$_{50}$ \\
		\hline
        GSPN \cite{yi2019gspn}&CVPR 19&17.7&30.6\\
        3D-SIS \cite{hou20193d}&CVPR 21&22.5&40.2\\
        3D-MPA \cite{engelmann20203d}&CVPR 20&49.2&64.2\\
        SoftGroup \cite{vu2022softgroup}&CVPR 22&59.4&71.6\\
        Mask3D \cite{Schult23ICRA}&ICRA 23&56.2&70.2\\
        {\bf Ours$^*$}&-&{\bf 60.2}&{\bf 72.2}\\
		\hline
	\end{tabular}
 \end{center}
 \label{tab7}
 \vspace{-4mm}
\end{table}

In addition, we apply our method to the 3D object detection task and compare it with other 3D instance segmentation methods adapted for this task. The experimental results in Table \ref{tab7} demonstrate the effectiveness of our EipFormer.

\begin{figure*}[htpb]
    \begin{center}
        \includegraphics[width=1\linewidth]{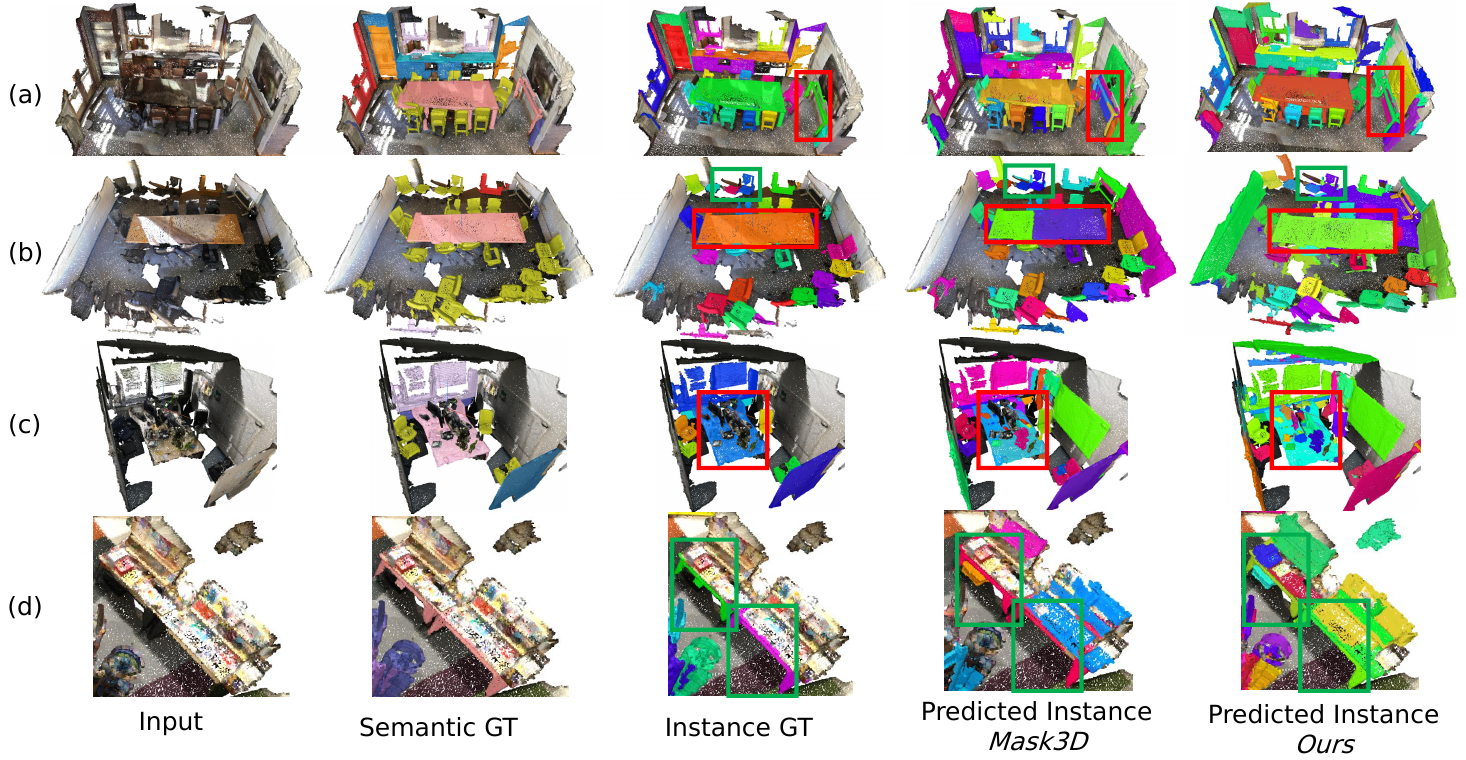}
    \end{center}
    \caption{Visual examples of 3D instance segmentation results on the validation set of ScanNetV2. Enlarge the image for improved visualization. The comparison with the Mask3D reveals that our proposed EipFormer effectively distinguishes neighboring instances (the green boxes in (b) and (d)) and yields better instance segmentation results (indicated by red boxes in (a), (b) and (d)).}
    \label{fig3}
\end{figure*}

\subsection{Qualitative Results}  \label{section4.3}
\begin{figure*}[htpb]
    \centering      
        \includegraphics[width=1\linewidth]{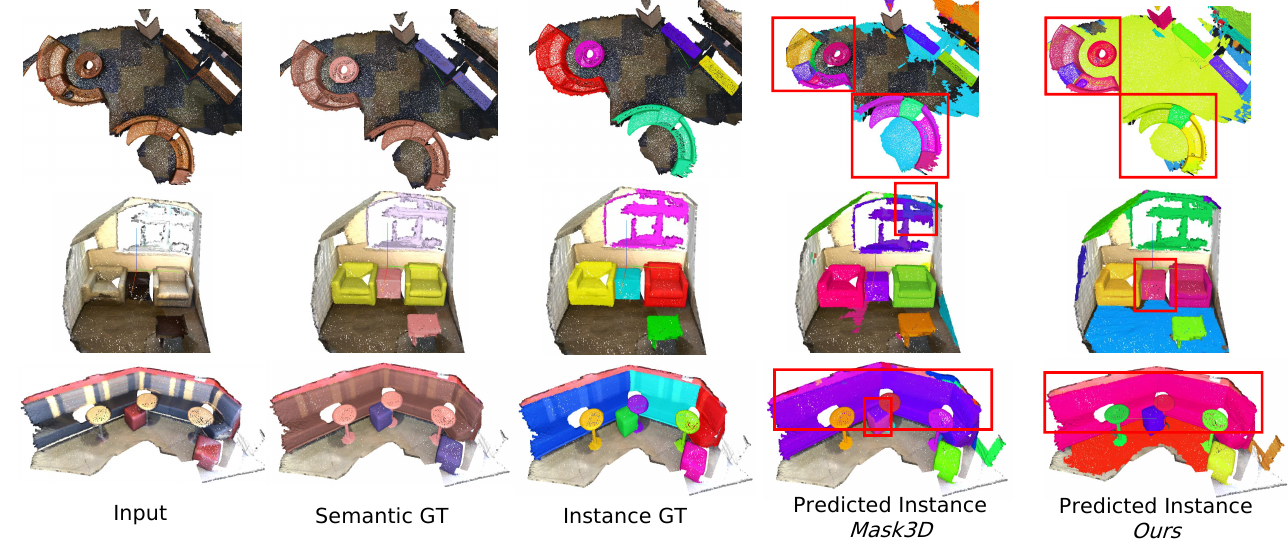}
    \caption{Failure cases on the validation set of ScanNetV2. Enlarge the image for improved visualization.}
    \label{fig2+}
\end{figure*}

Figure \ref{fig3} provides visual examples of 3D instance segmentation on the ScanNetV2 validation set.
The comparison with Mask3D reveals that our proposed EipFormer effectively distinguishes neighboring instances (the green boxes in Figure \ref{fig3} (b) and (d)) and yields better instance segmentation results (indicated by red boxes in Figure \ref{fig3} (a), (b), and (d)).
These results demonstrate the effectiveness of dual position embedding and progressive aggregation.
Furthermore, we have included several failure cases in Figure \ref{fig2+}.
In our view, these mis-predicted instance segmentation results more closely resemble reality than the annotated ground-truth labels.

\begin{table*}[htpb]
    \centering
	\caption{Ablation studies on various components. CAPS: The class-aware point sampling in Section \ref{section3.3}. O-PE: Using the original position embedding \text{Fourier}($\mathbf{C}$) to yield $\mathbf{E}$ in Eq. (\ref{eq3}). C-PE: Using the centralized position embedding \text{Fourier}($\mathbf{C+O}$) to yield $\mathbf{E}$ in Eq. (\ref{eq3}). Weighted-FPS: Weighted farthest point sampling.
 The \textit{Baseline} extracts instance features from embeddings of randomly sampled points.
It employs the original position encoding, the conventional farthest point sampling, and 2$L$-1 Transformer decoder layers with proposal matching.
The Fine and Merge stages are described in Section \ref{section3.2.2} and Section \ref{section3.2.3}.
	}
	\label{tab8}
	\begin{tabular}{ccccc cc|lll } 
		\hline 
		& {\bf CAPS} & {\bf O-PE} & {\bf C-PE} & {\bf Fine} & {\bf Merge} & {\bf Weighted-FPS}  & {\bf AP} & {\bf AP}$_{50}$ & {\bf AP}$_{25}$ \\ 
		\hline
		\textit{Baseline}&&$\checkmark$&&&&&51.2&69.8&78.7\\
		\hline
        \textit{V$_1$}&$\checkmark$&$\checkmark$&&&&&52.8&70.9&79.6\\
        \textit{V$_2$}&$\checkmark$&&$\checkmark$&&&&52.1&71.1&81.2\\
        \textit{V$_3$}&$\checkmark$&$\checkmark$&$\checkmark$&&&&54.1&72.0&81.5\\
        \textit{V$_4$}&$\checkmark$&$\checkmark$&$\checkmark$&$\checkmark$&&&47.6&66.8&78.8\\
        \textit{V$_5$}&$\checkmark$&$\checkmark$&$\checkmark$&&$\checkmark$&&54.3&72.4&81.8\\
        \textit{V$_6$}&$\checkmark$&$\checkmark$&$\checkmark$&$\checkmark$&$\checkmark$&&56.1&73.9&82.4\\
        \textit{V$_7$}&$\checkmark$&$\checkmark$&$\checkmark$&$\checkmark$&$\checkmark$&$\checkmark$&{\bf 56.9$^{+5.7}$}&{\bf 74.6$^{+4.8}$}&{\bf 82.5$^{+3.8}$}\\
		\hline
	\end{tabular}
\end{table*}

\subsection{Ablation Study}  \label{section4.4}
Ablation studies are conducted on the ScanNetV2 dataset. In Sections \ref{section4.4.1} $\sim$ \ref{section4.4.4}, we illustrate the influence of each proposed component on instance segmentation performance. Section \ref{section4.4.5} investigates the impact of query position embedding in the Fine stage on prediction performance. Furthermore, the effects of center offsets and loss functions supervising the weighted farthest point sampling are explored in Sections \ref{section4.4.6} and \ref{section4.4.7}, respectively. Additionally, in Section \ref{section4.4.8}, we evaluate the impact of various feature backbones on model performance.

Table \ref{tab8} presents results obtained through the gradual integration of various components into the baseline model. 
The \textit{Baseline} extracts instance features from embeddings of randomly sampled points. It employs the original position encoding, the conventional farthest point sampling, and $2L-1$ Transformer decoder layers with proposal matching.

\subsubsection{Impact of class-aware point sampling}\label{section4.4.1}
Instead of aggregating instance features using randomly sampled voxels, we employ class-aware point sampling. By comparing the results of \textit{V$_1$} and \textit{Baseline}, we can conclude that class-aware point sampling performs better than random sampling. Incorporating class-aware point sampling leads to a notable improvement in the AP/AP$_{50}$ metrics, from 51.2/69.8 to 52.8/70.9.

\subsubsection{Effect of dual position embedding}\label{section4.4.2}
The dual position embedding module combines the original position embedding and the centralized position embedding, as expressed in Eq. (\ref{eq3}). 
In variant \textit{V$_2$}, we replace the original position embedding with the centralized position embedding, resulting in improved performance for instance segmentation -- from 70.9/79.6 to 71.1/81.2 on AP$_{50}$/AP$_{25}$ metrics.
Furthermore, incorporating both the original and centralized position embeddings in \textit{V$_{3}$} leads to an additional increase -- from 52.1/71.1/81.2 to 54.1/72.0/81.5 on AP/AP$_{50}$/AP$_{25}$ metrics.
Namely, dual position embedding shows superior performance compared to single position embedding methods.

\begin{figure*}[htpb]
    \centering      
        \includegraphics[width=0.9\linewidth]{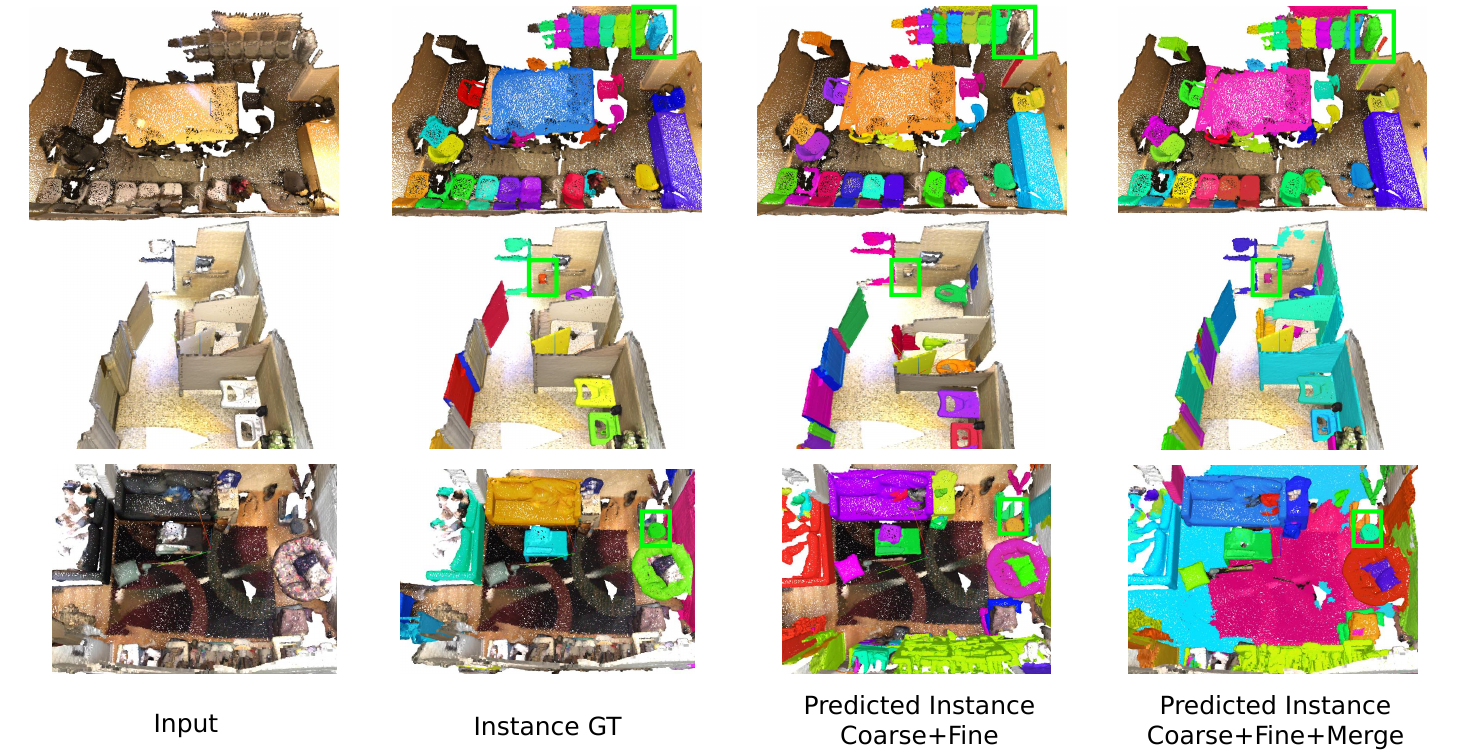}
    \caption{Visualization of omitted instances. The Fine stage may omit several instances (the green boxes), which will be resolved by incorporating `Merge'.}
    \label{figsuo1}
\end{figure*}

\subsubsection{Progressive aggregation}\label{section4.4.3}
\begin{figure*}[htpb]
    \centering      
        \includegraphics[width=0.9\linewidth]{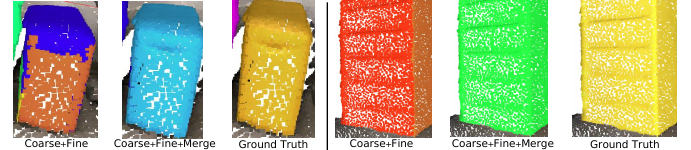}
    \caption{Visualization of fragmented instances. The Fine stage may yield several fragmented instances, which will be resolved by incorporating `Merge'.}
    \label{figsuo2}
\end{figure*}

\textit{V$_4$} presents the results of fusing the Fine stage in progressive aggregation. Specifically, we refine the instance positions using Eq. (\ref{eq9}) and Eq. (\ref{eq92}), and the instance features using center matching in Eq. (\ref{eq10}). It is evident that \textit{V$_4$} significantly degrades the prediction performance, $i.e.$, a decline of 6.5 in AP metric compared to \textit{V$_3$}.
As we have mentioned, the Fine stage concentrates on aggregating features of points close to instance centers, which may omit several instances and produce fragmented instances.
The visualization of several omitted and fragmented instances are shown in Figure \ref{figsuo1} and Figure \ref{figsuo2}, respectively.
\textit{V$_6$} integrates the Merge stage, improving the instance segmentation performance from 54.1/72.0/81.5 to 56.1/73.9/82.4 on AP/AP$_{50}$/AP$_{25}$ metrics. This highlights the effectiveness of the proposed progressive aggregation.
Notably, incorporating the Merge stage (\textit{V$_5$}) only leads to a slight improvement than the performance of \textit{V$_4$}.
This observation indicates that refining the instance positions is beneficial for yielding better instance segmentation results.

\subsubsection{Weighted farthest point sampling}\label{section4.4.4}
Replacing the farthest point sampling technique in \textit{V$_6$} with the weighted variant contributes to the instance segmentation performance. \textit{V$_7$} demonstrates optimal performance across all metrics, achieving a notable increase of +5.7/4.8/3.8 in AP/AP$_{50}$/AP$_{25}$ metrics compared to the \textit{Baseline}.

\subsubsection{Type of query position embedding in Eq. (\ref{eq92})}\label{section4.4.5}
Section \ref{section4.4.3} has demonstrated the effectiveness of the Fine stage. We then investigate the influence of various query position embeddings $\mathbf{Q_e}$ in the Fine stage on instance segmentation results. $\mathbf{Q_e}$ plays a crucial role in guiding the cross-attention and self-attention mechanisms during the feature aggregation process.
We distinguish between two types of $\mathbf{Q_e}$: $\mathbf{Q}_e^{0}$ and $\mathbf{Q}_e^{avg}$. $\mathbf{Q}_e^{0}$ represents the Fourier encoding of initially selected representative points, while $\mathbf{Q}_e^{avg}$ applies the Fourier encoding to the averaged coordinates $\mathbf{C}_i^{avg}$ in Eq. (\ref{eq9}). Furthermore, $\mathbf{C}$ in Eq. (\ref{eq9}) can be replaced with $\mathbf{C+O}$.
Table \ref{tab9} displays the detailed experimental results. We observe that $\mathbf{Q}_e^{avg}$ with $\mathbf{C}$ outperforms the other variants.

\begin{table}[tpb]
 \centering
 	\caption{Ablation studies on various types of query position embeddings $\mathbf{Q_e}$.
  We distinguish two types of $\mathbf{Q_e}$: $\mathbf{Q}_e^{0}$ and $\mathbf{Q}_e^{avg}$.
$\mathbf{Q}_e^{0}$ means the Fourier encoding of initially selected representative points, while $\mathbf{Q}_e^{avg}$ applies the Fourier encoding to the averaged coordinates $\mathbf{C}_i^{avg}$ in Eq. (\ref{eq9}).
Furthermore, $\mathbf{C}$ in Eq. (\ref{eq9}) can be replaced with $\mathbf{C+O}$.
  }
 	\label{tab9}
	\begin{tabular}{lc|ccc } 
		\hline 
        \multicolumn{2}{c|}{{\bf $\mathbf{Q_e}$}}&{\bf AP}&{\bf AP}$_{50}$&{\bf AP}$_{25}$\\
		\hline
        $\mathbf{Q}_e^{0}$ &-&55.4&73.3&{\bf 82.6}\\
        \hline
        \multirow{2}{*}{$\mathbf{Q}_e^{avg}$}
        &$\mathbf{C}$&{\bf 56.9}&{\bf 74.6}&{82.5}\\
        & $\mathbf{C+O}$&53.8&72.0&81.8\\
		\hline
	\end{tabular}
 \end{table}

\begin{table}[tpb]
    \centering
	\caption{Ablation studies on the type of ground truth centers in Eq. (\ref{eq4}). `FPS' represents the farthest point sampling.}
	\label{tab10}
	\begin{tabular}{cc|ccc } 
		\hline
        \multicolumn{2}{c|}{\bf GT Centers}&{\bf AP}&{\bf AP}$_{50}$&{\bf AP}$_{25}$\\
        \hline
        \multicolumn{2}{c}{FPS} &56.1&73.9&82.4\\
		\hline
        \multirow{3}{*}{Weighted-FPS}&       
        Average Center&{\bf 56.9}&{\bf 74.6}&82.5\\
        &Median Center&55.4&73.9&{\bf 82.9}\\
        &Box Center&52.6&71.2&80.7\\
		\hline
	\end{tabular}
\end{table}

\begin{table}[tpb]
    \centering
 	\caption{Ablation studies on various losses in the weighted farthest point sampling. 
  $\mathcal{L}_{fore}$ and $\mathcal{L}_{back}$ represent the foreground and background losses, respectively. The Semantic loss denotes the cross-entropy loss for predicting the semantic label of voxels within 3D point clouds.}
	\label{tab11}
	\begin{tabular}{ccc|ccc } 
		\hline 
        $\mathcal{L}_{fore}$ & $\mathcal{L}_{back}$ & {\bf Semantic Loss}& {\bf AP}&{\bf AP}$_{50}$&{\bf AP}$_{25}$\\
		\hline
        $\checkmark$&&&54.5&72.7&82.3\\
        &$\checkmark$&&{56.8}&73.8&81.8\\
        $\checkmark$&$\checkmark$&&{\bf  56.9}&{\bf 74.6}&82.5\\
        $\checkmark$&$\checkmark$&$\checkmark$ &55.4&73.1&{\bf 82.7}\\
		\hline
	\end{tabular}
\end{table}

\subsubsection{Impact of $o^*$ in Eq. (\ref{eq4})}\label{section4.4.6}
Section \ref{section4.4.4} has shown the effect of weighted farthest point sampling. Next, we explore the influence of different ground truth centers guiding the weighting factor on instance segmentation. These centers encompass the average center, median center, and box center. The comparative results are summarized in Table \ref{tab10}.
Notably, the weighted farthest point sampling supervised by average centers achieves optimal performance. 
This observation suggests that utilizing the average center can better aggregate points into instances.
Additionally, within the 3D context, using bounding boxes to describe instances may encompass redundant regions. Consequently, in comparison with alternative methods, the ground truth offsets of points to their box centers exhibit significantly higher values.

\subsubsection{Losses in weighted farthest point sampling} \label{section4.4.7}
Building on the results in Section \ref{section4.4.6}, we choose average centers as ground truth centers and proceed to analyze the loss functions in weighted farthest point sampling. 
This analysis involves three distinct loss types: the foreground loss $\mathcal{L}_{fore}$ in Eq. (\ref{eq4}), the background loss $\mathcal{L}_{back}$ in Eq. (\ref{eq5}), and the cross-entropy loss for predicting the semantic label of each voxel. The experimental results are presented in Table \ref{tab11}.
From these results, we observe that weighted farthest point sampling with $\mathcal{L}_{back}$ exhibits superior performance on AP/AP$_{50}$ metrics compared to that with $\mathcal{L}_{fore}$. Furthermore, the combination of both losses, $\mathcal{L}_{fore}$ and $\mathcal{L}_{back}$, brings further performance improvement.
Moreover, when integrating semantic confidence with the predicted $\mathbf{W}$ in Eqs. (\ref{eq4}) and (\ref{eq5}), a reduction in performance becomes apparent. In our opinion, this may be attributed to imprecise semantic predictions.

\subsubsection{Various feature backbones} \label{section4.4.8}
On the ScanNet dataset, we employ the Minkowski Res16UNet34C as the feature backbone and set the number of sampled points mentioned in Section \ref{section3.3} to 12,800. 
Additionally, we conduct a comparative experiment on this dataset, employing the Minkowski Res16UNet101 as the feature backbone while adjusting the number of sampled points to 6,400 accordingly. All experiments are performed on a single GeForce RTX 3090 GPU. It is worth noting that maintaining the sampled points at 12,800 with Minkowski Res16UNet101 as the feature backbone results in `out-of-memory'.
Detailed results are presented in Table \ref{tab12}.
The results suggest that using a deeper architecture has a marginal effect on instance segmentation performance. For example, there is a slight improvement from 74.6 to 74.8 in terms of AP$_{50}$.

\begin{table}[tpb]
    \centering
	\caption{Ablation studies on various feature backbones.}
	\label{tab12}
	\begin{tabular}{l|ccc } 
		\hline
        {\bf Backbones}&{\bf AP}&{\bf AP}$_{50}$&{\bf AP}$_{25}$\\
        \hline
        Res16UNet34C &{\bf 56.9}&74.6&82.5\\
        Res16UNet101&56.7&{\bf 74.8}&{\bf 82.9}\\
		\hline
	\end{tabular}
\end{table}

\begin{table}[tpb]
 	\caption{Ablation studies on the impact of instance positions. \textit{Baseline} represents the variant $V_5$ in Table \ref{tab4}. `CM' means the center matching. The `Pred center' denotes the centrally shifted points predicted by Eq. (\ref{eq1}) and Eq. (\ref{eq2}). The `Agg averaging' signifies the aggression averaging in Eq. (\ref{eq9}). }
	\label{tab13}
 \centering
	\begin{tabular}{l|ccc } 
		\hline
        &{\bf AP}&{\bf AP}$_{50}$&{\bf AP}$_{25}$\\
        \hline
        \textit{Baseline} &54.3&72.4&81.8\\
        GT center \& CM&{\bf 66.1}&{\bf 81.1}&{\bf 85.9}\\
        Pred center \& CM&45.6&64.9&77.8\\
        Agg averaging \& CM&{56.1}&{73.9}&{82.4}\\
		\hline
	\end{tabular}
\end{table}

\section{Other attempts}\label{section4.5}
In this section, we elaborate on our experimental endeavors, providing details of our attempts and their outcomes.

\subsubsection{Impact of instance positions} 
Center prediction or farthest point sampling has been widely used to generate fixed instance positions as representative points in previous approaches. 
However, a challenge arises as the majority of points sampled through farthest point sampling often originate from backgrounds or belong to large instances. Simultaneously, achieving precise offset regression for all points through the center offset prediction branch poses difficulties.
To illustrate these issues, we conduct a series of experiments, and show their results in Table \ref{tab13}.

Firstly, we employ ground truth instance centers to generate position embeddings for initial instance queries. Surprisingly, this does not bring about notable performance changes. Building upon this, we replace proposal matching with the proposed center matching, resulting in a significant performance enhancement. For instance, it achieves an absolute improvement of 11.8/8.7/4.1 on AP/AP$_{50}$/AP$_{25}$ metrics compared to the \textit{Baseline}. This finding highlights the importance of establishing a one-to-one correspondence between instance positions and instance targets.

Subsequently, we substitute the ground truth center with the predicted centers ($i.e.$, the centrally shifted points) and observe a decline of 20.5/16.2/8.1 on AP/AP$_{50}$/AP$_{25}$ metrics. This decline can be attributed to the considerable deviation between the predicted and ground truth instance centers. In our method, we employ weighted farthest point sampling to enhance initial instance positions, coupled with aggregation averaging to obtain refined instance positions. As observed, our proposed method outperforms other variants, except for the one with ground-truth centers.
Inspired by the results that using ground-truth centers leads to an absolute 10-point improvement on the AP metric over the state-of-the-art, we believe that enhancing the performance of predicting instance centers is a promising direction for future research.

\subsubsection{Residual connections}
In the Fine stage, we establish a specific connection between instance positions and instance targets, enabling residual connections between various Transformer decoder layers. 
However, these residual connections have limited impact on the instance segmentation results.

\subsubsection{Integrate $\mathcal{L}_{cm}$ in Eq. (\ref{eq10}) to $\mathcal{L}_{all}$ in Eq. (\ref{eq11})}
In this work, $\mathcal{L}_{cm}$ is solely utilized to assign instance targets for predicted instance segmentation results. Subsequently, we incorporate this loss into $\mathcal{L}_{all}$ for model training. The experimental results are provided in Table \ref{tab14}. 
It is evident that incorporating $\mathcal{L}_{cm}$ does not yield improvement in instance segmentation performance.

\begin{table}[t]
    \centering
    \caption{Ablation studies to assess the impact of $\mathcal{L}_{cm}$ on model training.}
    \label{tab14}
    \begin{tabular}{ccccc}
    \hline
$\mathcal{L}_{ac}$&$\mathcal{L}_{cm}$&{\bf Merge}& {\bf AP}&{\bf AP$_{50}$}\\
\hline
$\checkmark$&&&54.1&72.0\\
$\checkmark$&$\checkmark$&&52.8&71.7\\
$\checkmark$&$\checkmark$&$\checkmark$&54.7&72.9\\
$\checkmark$&&$\checkmark$&{\bf 56.1}&{\bf 73.9}\\
\hline
    \end{tabular}
\end{table}

\section{Conclusion}
In this study, we introduce EipFormer, a Transformer-based bottom-up approach designed for the challenging task of 3D instance segmentation. 
EipFormer utilizes progressive aggregation to refine instance query positions and features, incorporating weighted farthest point sampling, aggregation averaging, and center matching techniques. Additionally, dual position embedding is employed to distinguish neighboring instances by superposing both the original and centralized position embeddings.
To optimize memory consumption, we design a class-aware point sampling strategy, effectively aggregating instance features from sampled points. Extensive experiments conducted on various popular instance segmentation datasets demonstrate the superior performance of EipFormer compared to existing state-of-the-art methods.

\section*{Acknowledgment}
This work was supported by the National Natural Science Foundation of China \#U19B2039 and \#62276046.

\bibliographystyle{IEEEtran}
\bibliography{ref}

\begin{thebibliography}{10}
\providecommand{\url}[1]{#1}
\csname url@samestyle\endcsname
\providecommand{\newblock}{\relax}
\providecommand{\bibinfo}[2]{#2}
\providecommand{\BIBentrySTDinterwordspacing}{\spaceskip=0pt\relax}
\providecommand{\BIBentryALTinterwordstretchfactor}{4}
\providecommand{\BIBentryALTinterwordspacing}{\spaceskip=\fontdimen2\font plus
\BIBentryALTinterwordstretchfactor\fontdimen3\font minus \fontdimen4\font\relax}
\providecommand{\BIBforeignlanguage}[2]{{%
\expandafter\ifx\csname l@#1\endcsname\relax
\typeout{** WARNING: IEEEtran.bst: No hyphenation pattern has been}%
\typeout{** loaded for the language `#1'. Using the pattern for}%
\typeout{** the default language instead.}%
\else
\language=\csname l@#1\endcsname
\fi
#2}}
\providecommand{\BIBdecl}{\relax}
\BIBdecl

\bibitem{wei2020mitoem}
D.~Wei, Z.~Lin, D.~Franco-Barranco, N.~Wendt, X.~Liu, W.~Yin, X.~Huang, A.~Gupta, W.-D. Jang, X.~Wang \emph{et~al.}, ``Mitoem dataset: Large-scale 3d mitochondria instance segmentation from em images,'' in \emph{International Conference on Medical Image Computing and Computer-Assisted Intervention}.\hskip 1em plus 0.5em minus 0.4em\relax Springer, 2020, pp. 66--76.

\bibitem{chen2022stpls3d}
M.~Chen, Q.~Hu, Z.~Yu, H.~Thomas, A.~Feng, Y.~Hou, K.~McCullough, F.~Ren, and L.~Soibelman, ``Stpls3d: A large-scale synthetic and real aerial photogrammetry 3d point cloud dataset,'' \emph{arXiv preprint arXiv:2203.09065}, 2022.

\bibitem{hou2021exploring}
J.~Hou, B.~Graham, M.~Nie{\ss}ner, and S.~Xie, ``Exploring data-efficient 3d scene understanding with contrastive scene contexts,'' in \emph{IEEE/CVF Conference on Computer Vision and Pattern Recognition}, 2021, pp. 15\,587--15\,597.

\bibitem{chen20224dcontrast}
Y.~Chen, M.~Nie{\ss}ner, and A.~Dai, ``4dcontrast: Contrastive learning with dynamic correspondences for 3d scene understanding,'' in \emph{European Conference on Computer Vision}.\hskip 1em plus 0.5em minus 0.4em\relax Springer, 2022, pp. 543--560.

\bibitem{jaritz2019multi}
M.~Jaritz, J.~Gu, and H.~Su, ``Multi-view pointnet for 3d scene understanding,'' in \emph{IEEE/CVF International Conference on Computer Vision Workshops}, 2019, pp. 0--0.

\bibitem{zhang2021holistic}
C.~Zhang, Z.~Cui, Y.~Zhang, B.~Zeng, M.~Pollefeys, and S.~Liu, ``Holistic 3d scene understanding from a single image with implicit representation,'' in \emph{IEEE/CVF Conference on Computer Vision and Pattern Recognition}, 2021, pp. 8833--8842.

\bibitem{jiang2020end}
H.~Jiang, F.~Yan, J.~Cai, J.~Zheng, and J.~Xiao, ``End-to-end 3d point cloud instance segmentation without detection,'' in \emph{IEEE/CVF Conference on Computer Vision and Pattern Recognition}, 2020, pp. 12\,796--12\,805.

\bibitem{wen2020cf}
X.~Wen, Z.~Han, G.~Youk, and Y.-S. Liu, ``Cf-sis: Semantic-instance segmentation of 3d point clouds by context fusion with self-attention,'' in \emph{ACM International Conference on Multimedia}, 2020, pp. 1661--1669.

\bibitem{zhou2020joint}
D.~Zhou, J.~Fang, X.~Song, L.~Liu, J.~Yin, Y.~Dai, H.~Li, and R.~Yang, ``Joint 3d instance segmentation and object detection for autonomous driving,'' in \emph{IEEE/CVF Conference on Computer Vision and Pattern Recognition}, 2020, pp. 1839--1849.

\bibitem{wang2021solo}
X.~Wang, R.~Zhang, C.~Shen, T.~Kong, and L.~Li, ``Solo: A simple framework for instance segmentation,'' \emph{IEEE transactions on pattern analysis and machine intelligence}, vol.~44, no.~11, pp. 8587--8601, 2021.

\bibitem{xie2021unseen}
C.~Xie, Y.~Xiang, A.~Mousavian, and D.~Fox, ``Unseen object instance segmentation for robotic environments,'' \emph{IEEE Transactions on Robotics}, vol.~37, no.~5, pp. 1343--1359, 2021.

\bibitem{zanjani2019mask}
F.~G. Zanjani, D.~A. Moin, F.~Claessen, T.~Cherici, S.~Parinussa, A.~Pourtaherian, S.~Zinger, and P.~H. de~With, ``Mask-mcnet: Instance segmentation in 3d point cloud of intra-oral scans,'' in \emph{Medical Image Computing and Computer Assisted Intervention}.\hskip 1em plus 0.5em minus 0.4em\relax Springer, 2019, pp. 128--136.

\bibitem{hu2018semantic}
S.-M. Hu, J.-X. Cai, and Y.-K. Lai, ``Semantic labeling and instance segmentation of 3d point clouds using patch context analysis and multiscale processing,'' \emph{IEEE transactions on visualization and computer graphics}, vol.~26, no.~7, pp. 2485--2498, 2018.

\bibitem{li2022joint}
Y.~Li, J.~Cai, Q.~Zhou, and H.~Lu, ``Joint semantic-instance segmentation method for intelligent transportation system,'' \emph{IEEE Transactions on Intelligent Transportation Systems}, 2022.

\bibitem{Schult23ICRA}
J.~Schult, F.~Engelmann, A.~Hermans, O.~Litany, S.~Tang, and B.~Leibe, ``{Mask3D for 3D Semantic Instance Segmentation},'' 2023.

\bibitem{chen20223}
J.~Chen, Y.~Xu, S.~Lu, R.~Liang, and L.~Nan, ``3-d instance segmentation of mvs buildings,'' \emph{IEEE Transactions on Geoscience and Remote Sensing}, vol.~60, pp. 1--14, 2022.

\bibitem{yang2019learning}
B.~Yang, J.~Wang, R.~Clark, Q.~Hu, S.~Wang, A.~Markham, and N.~Trigoni, ``Learning object bounding boxes for 3d instance segmentation on point clouds,'' \emph{Advances in Neural Information Processing Systems}, vol.~32, 2019.

\bibitem{liu2020learning}
S.-H. Liu, S.-Y. Yu, S.-C. Wu, H.-T. Chen, and T.-L. Liu, ``Learning gaussian instance segmentation in point clouds,'' \emph{arXiv preprint arXiv:2007.09860}, 2020.

\bibitem{wang2018sgpn}
W.~Wang, R.~Yu, Q.~Huang, and U.~Neumann, ``Sgpn: Similarity group proposal network for 3d point cloud instance segmentation,'' in \emph{IEEE/CVF Conference on Computer Vision and Pattern Recognition}, 2018, pp. 2569--2578.

\bibitem{hou20193d}
J.~Hou, A.~Dai, and M.~Nie{\ss}ner, ``3d-sis: 3d semantic instance segmentation of rgb-d scans,'' in \emph{IEEE/CVF Conference on Computer Vision and Pattern Recognition}, 2019, pp. 4421--4430.

\bibitem{engelmann20203d}
F.~Engelmann, M.~Bokeloh, A.~Fathi, B.~Leibe, and M.~Nie{\ss}ner, ``3d-mpa: Multi-proposal aggregation for 3d semantic instance segmentation,'' in \emph{IEEE/CVF Conference on Computer Vision and Pattern Recognition}, 2020, pp. 9031--9040.

\bibitem{chen2021hierarchical}
S.~Chen, J.~Fang, Q.~Zhang, W.~Liu, and X.~Wang, ``Hierarchical aggregation for 3d instance segmentation,'' in \emph{IEEE/CVF International Conference on Computer Vision}, 2021, pp. 15\,467--15\,476.

\bibitem{wang2019associatively}
X.~Wang, S.~Liu, X.~Shen, C.~Shen, and J.~Jia, ``Associatively segmenting instances and semantics in point clouds,'' in \emph{IEEE/CVF Conference on Computer Vision and Pattern Recognition}, 2019, pp. 4096--4105.

\bibitem{jiang2020pointgroup}
L.~Jiang, H.~Zhao, S.~Shi, S.~Liu, C.-W. Fu, and J.~Jia, ``Pointgroup: Dual-set point grouping for 3d instance segmentation,'' in \emph{IEEE/CVF Conference on Computer Vision and Pattern Recognition}, 2020, pp. 4867--4876.

\bibitem{cciccek20163d}
{\"O}.~{\c{C}}i{\c{c}}ek, A.~Abdulkadir, S.~S. Lienkamp, T.~Brox, and O.~Ronneberger, ``3d u-net: learning dense volumetric segmentation from sparse annotation,'' in \emph{Medical Image Computing and Computer-Assisted Intervention}.\hskip 1em plus 0.5em minus 0.4em\relax Springer, 2016, pp. 424--432.

\bibitem{graham20183d}
B.~Graham, M.~Engelcke, and L.~Van Der~Maaten, ``3d semantic segmentation with submanifold sparse convolutional networks,'' in \emph{IEEE/CVF Conference on Computer Vision and Pattern Recognition}, 2018, pp. 9224--9232.

\bibitem{liao2021point}
Y.~Liao, H.~Zhu, Y.~Zhang, C.~Ye, T.~Chen, and J.~Fan, ``Point cloud instance segmentation with semi-supervised bounding-box mining,'' \emph{IEEE Transactions on Pattern Analysis and Machine Intelligence}, vol.~44, no.~12, pp. 10\,159--10\,170, 2021.

\bibitem{qi2019deep}
C.~R. Qi, O.~Litany, K.~He, and L.~J. Guibas, ``Deep hough voting for 3d object detection in point clouds,'' in \emph{IEEE/CVF International Conference on Computer Vision}, 2019, pp. 9277--9286.

\bibitem{SPFormer}
S.~Jiahao, Q.~Chunmei, T.~Junpeng, and X.~Xiangmin, ``Superpoint transformer for 3d scene instance segmentation,'' in \emph{AAAI Conference on Artificial Intelligence}, 2023.

\bibitem{ngo2023isbnet}
T.~D. Ngo, B.-S. Hua, and K.~Nguyen, ``Isbnet: a 3d point cloud instance segmentation network with instance-aware sampling and box-aware dynamic convolution,'' in \emph{IEEE/CVF Conference on Computer Vision and Pattern Recognition}, 2023, pp. 13\,550--13\,559.

\bibitem{yin2021bridging}
C.~Yin, J.~Tang, T.~Yuan, Z.~Xu, and Y.~Wang, ``Bridging the gap between semantic segmentation and instance segmentation,'' \emph{IEEE Transactions on Multimedia}, vol.~24, pp. 4183--4196, 2021.

\bibitem{he2023prototype}
S.~He, X.~Jiang, W.~Jiang, and H.~Ding, ``Prototype adaption and projection for few-and zero-shot 3d point cloud semantic segmentation,'' \emph{IEEE Transactions on Image Processing}, 2023.

\bibitem{cheng2021net}
S.~Cheng, X.~Chen, X.~He, Z.~Liu, and X.~Bai, ``Pra-net: Point relation-aware network for 3d point cloud analysis,'' \emph{IEEE Transactions on Image Processing}, vol.~30, pp. 4436--4448, 2021.

\bibitem{gu20193d}
S.~Gu, J.~Hou, H.~Zeng, H.~Yuan, and K.-K. Ma, ``3d point cloud attribute compression using geometry-guided sparse representation,'' \emph{IEEE Transactions on Image Processing}, vol.~29, pp. 796--808, 2019.

\bibitem{qi2017pointnet}
C.~R. Qi, H.~Su, K.~Mo, and L.~J. Guibas, ``Pointnet: Deep learning on point sets for 3d classification and segmentation,'' in \emph{IEEE/CVF Conference on Computer Vision and Pattern Recognition}, 2017, pp. 652--660.

\bibitem{qi2017pointnet++}
C.~R. Qi, L.~Yi, H.~Su, and L.~J. Guibas, ``Pointnet++: Deep hierarchical feature learning on point sets in a metric space,'' \emph{Advances in Neural Information Processing Systems}, vol.~30, 2017.

\bibitem{li2018pointcnn}
Y.~Li, R.~Bu, M.~Sun, W.~Wu, X.~Di, and B.~Chen, ``Pointcnn: Convolution on x-transformed points,'' \emph{Advances in Neural Information Processing Systems}, vol.~31, 2018.

\bibitem{ronneberger2015u}
O.~Ronneberger, P.~Fischer, and T.~Brox, ``U-net: Convolutional networks for biomedical image segmentation,'' in \emph{Medical Image Computing and Computer-Assisted Intervention}.\hskip 1em plus 0.5em minus 0.4em\relax Springer, 2015, pp. 234--241.

\bibitem{carreira2017quo}
J.~Carreira and A.~Zisserman, ``Quo vadis, action recognition? a new model and the kinetics dataset,'' in \emph{IEEE/CVF Conference on Computer Vision and Pattern Recognition}, 2017, pp. 6299--6308.

\bibitem{he2016deep}
K.~He, X.~Zhang, S.~Ren, and J.~Sun, ``Deep residual learning for image recognition,'' in \emph{IEEE/CVF Conference on Computer Vision and Pattern Recognition}, 2016, pp. 770--778.

\bibitem{simonyan2014very}
K.~Simonyan and A.~Zisserman, ``Very deep convolutional networks for large-scale image recognition,'' \emph{arXiv preprint arXiv:1409.1556}, 2014.

\bibitem{yan2018second}
Y.~Yan, Y.~Mao, and B.~Li, ``Second: Sparsely embedded convolutional detection,'' \emph{Sensors}, vol.~18, no.~10, p. 3337, 2018.

\bibitem{yi2019gspn}
L.~Yi, W.~Zhao, H.~Wang, M.~Sung, and L.~J. Guibas, ``Gspn: Generative shape proposal network for 3d instance segmentation in point cloud,'' in \emph{IEEE/CVF Conference on Computer Vision and Pattern Recognition}, 2019, pp. 3947--3956.

\bibitem{zhao2020jsnet}
L.~Zhao and W.~Tao, ``Jsnet: Joint instance and semantic segmentation of 3d point clouds,'' in \emph{AAAI Conference on Artificial Intelligence}, vol.~34, no.~07, 2020, pp. 12\,951--12\,958.

\bibitem{liang2021instance}
Z.~Liang, Z.~Li, S.~Xu, M.~Tan, and K.~Jia, ``Instance segmentation in 3d scenes using semantic superpoint tree networks,'' in \emph{IEEE/CVF International Conference on Computer Vision}, 2021, pp. 2783--2792.

\bibitem{vu2022softgroup}
T.~Vu, K.~Kim, T.~M. Luu, T.~Nguyen, and C.~D. Yoo, ``Softgroup for 3d instance segmentation on point clouds,'' in \emph{IEEE/CVF Conference on Computer Vision and Pattern Recognition}, 2022, pp. 2708--2717.

\bibitem{vu2022softgroup++}
T.~Vu, K.~Kim, T.~M. Luu, T.~Nguyen, J.~Kim, and C.~D. Yoo, ``Softgroup++: Scalable 3d instance segmentation with octree pyramid grouping,'' \emph{IEEE Transactions on Pattern Analysis and Machine Intelligence}, 2023.

\bibitem{choy20194d}
C.~Choy, J.~Gwak, and S.~Savarese, ``4d spatio-temporal convnets: Minkowski convolutional neural networks,'' in \emph{IEEE/CVF Conference on Computer Vision and Pattern Recognition}, 2019, pp. 3075--3084.

\bibitem{tancik2020fourier}
M.~Tancik, P.~Srinivasan, B.~Mildenhall, S.~Fridovich-Keil, N.~Raghavan, U.~Singhal, R.~Ramamoorthi, J.~Barron, and R.~Ng, ``Fourier features let networks learn high frequency functions in low dimensional domains,'' \emph{Advances in Neural Information Processing Systems}, vol.~33, pp. 7537--7547, 2020.

\bibitem{deng2018learning}
R.~Deng, C.~Shen, S.~Liu, H.~Wang, and X.~Liu, ``Learning to predict crisp boundaries,'' in \emph{European Conference on Computer Vision}, 2018, pp. 562--578.

\bibitem{cheng2022masked}
B.~Cheng, I.~Misra, A.~G. Schwing, A.~Kirillov, and R.~Girdhar, ``Masked-attention mask transformer for universal image segmentation,'' in \emph{IEEE/CVF Conference on Computer Vision and Pattern Recognition}, 2022, pp. 1290--1299.

\bibitem{armeni20163d}
I.~Armeni, O.~Sener, A.~R. Zamir, H.~Jiang, I.~Brilakis, M.~Fischer, and S.~Savarese, ``3d semantic parsing of large-scale indoor spaces,'' in \emph{IEEE/CVF Conference on Computer Vision and Pattern Recognition}, 2016, pp. 1534--1543.

\bibitem{dai2017scannet}
A.~Dai, A.~X. Chang, M.~Savva, M.~Halber, T.~Funkhouser, and M.~Nie{\ss}ner, ``Scannet: Richly-annotated 3d reconstructions of indoor scenes,'' in \emph{IEEE/CVF Conference on Computer Vision and Pattern Recognition}, 2017, pp. 5828--5839.

\bibitem{zhao2022divide}
W.~Zhao, Y.~Yan, C.~Yang, J.~Ye, X.~Yang, and K.~Huang, ``Divide and conquer: 3d point cloud instance segmentation with point-wise binarization,'' 2022.

\bibitem{nekrasov2021mix3d}
A.~Nekrasov, J.~Schult, O.~Litany, B.~Leibe, and F.~Engelmann, ``Mix3d: Out-of-context data augmentation for 3d scenes,'' in \emph{International Conference on 3D Vision (3DV)}.\hskip 1em plus 0.5em minus 0.4em\relax IEEE, 2021, pp. 116--125.

\bibitem{han2020occuseg}
L.~Han, T.~Zheng, L.~Xu, and L.~Fang, ``Occuseg: Occupancy-aware 3d instance segmentation,'' in \emph{IEEE/CVF Conference on Computer Vision and Pattern Recognition}, 2020, pp. 2940--2949.

\bibitem{lahoud20193d}
J.~Lahoud, B.~Ghanem, M.~Pollefeys, and M.~R. Oswald, ``3d instance segmentation via multi-task metric learning,'' in \emph{IEEE/CVF International Conference on Computer Vision}, 2019, pp. 9256--9266.

\bibitem{chu2022twist}
R.~Chu, X.~Ye, Z.~Liu, X.~Tan, X.~Qi, C.-W. Fu, and J.~Jia, ``Twist: Two-way inter-label self-training for semi-supervised 3d instance segmentation,'' in \emph{IEEE/CVF Conference on Computer Vision and Pattern Recognition}, 2022, pp. 1100--1109.

\bibitem{hui2022learning}
L.~Hui, L.~Tang, Y.~Shen, J.~Xie, and J.~Yang, ``Learning superpoint graph cut for 3d instance segmentation,'' \emph{Advances in Neural Information Processing Systems}, vol.~35, pp. 36\,804--36\,817, 2022.

\bibitem{dong2022learning}
S.~Dong, G.~Lin, and T.-Y. Hung, ``Learning regional purity for instance segmentation on 3d point clouds,'' in \emph{European Conference on Computer Vision}.\hskip 1em plus 0.5em minus 0.4em\relax Springer, 2022, pp. 56--72.

\bibitem{he2022pointinst3d}
T.~He, W.~Yin, C.~Shen, and A.~van~den Hengel, ``Pointinst3d: Segmenting 3d instances by points,'' in \emph{European Conference on Computer Vision}.\hskip 1em plus 0.5em minus 0.4em\relax Springer, 2022, pp. 286--302.

\bibitem{wu20223d}
Y.~Wu, M.~Shi, S.~Du, H.~Lu, Z.~Cao, and W.~Zhong, ``3d instances as 1d kernels,'' in \emph{European Conference on Computer Vision}.\hskip 1em plus 0.5em minus 0.4em\relax Springer, 2022, pp. 235--252.

\end{thebibliography}

\end{document}